\documentclass[runningheads]{llncs}

\usepackage{hyperref}
\usepackage{amsmath}
\usepackage[]{algorithm2e}
\usepackage{color, colortbl}

\usepackage{graphicx}
\usepackage{paralist}
\usepackage[shortlabels]{enumitem}
\usepackage[]{todonotes}
\presetkeys%
    {todonotes}%
    {inline,backgroundcolor=yellow}{}

\newcommand{\vecX}{\mathbf{x}}
\newcommand{\vecY}{\mathbf{y}}
\newcommand{\vecZ}{\mathbf{z}}

\definecolor{Gray}{gray}{0.9}

\begin{document}
\title{Gradient-based adversarial attacks on categorical sequence models via traversing an embedded world} % \thanks{Supported by Skoltech}
\titlerunning{Gradient-based adversarial attacks}
\author{Ivan Fursov\inst{1} \and
Alexey Zaytsev\inst{1} \and
Nikita Kluchnikov\inst{1} \and
Andrey Kravchenko\inst{2} \and
Evgeny Burnaev\inst{1}}

% \author{Anonymous authors}
%
% \authorrunning{I. Fursov et al.}
% \institute{Anonymous institute}

\institute{Skolkovo Institute of Science and Technology, Moscow, Russia
\email{\{ivan.fursov,a.zaytsev,n.kluchnikov,e.burnaev\}@skoltech.ru}\\
\and
DeepReason.ai, Oxford, UK, \\
\email{andrey.kravchenko@deepreason.ai}}
\maketitle              % typeset the header of the contribution
\begin{abstract}
  Deep learning models suffer from a phenomenon called adversarial attacks:
  we can apply minor changes to the model input to fool a classifier for a particular example.
  The literature mostly considers adversarial attacks on models with images and other structured inputs. However, the adversarial attacks for categorical sequences can also be harmful. Successful attacks for inputs in the form of  categorical sequences should address the following challenges:
  \begin{inparaenum}[\bfseries(1)] \item non-differentiability of the target function, \item constraints on transformations of initial sequences, and \item diversity of possible problems. \end{inparaenum}
  We handle these challenges using two black-box adversarial attacks.
  The first approach adopts a Monte-Carlo method and allows usage in any scenario, 
  the second approach uses a continuous relaxation of models and target metrics, and thus allows a usage of state-of-the-art methods for adversarial attacks with little additional effort.
  Results for money transactions, medical fraud, and NLP datasets suggest that the proposed methods generate reasonable adversarial sequences that are close to original ones, but fool machine learning models.

\end{abstract}

\begin{figure}[!ht]
    \centering
  \includegraphics[width=0.7\textwidth]{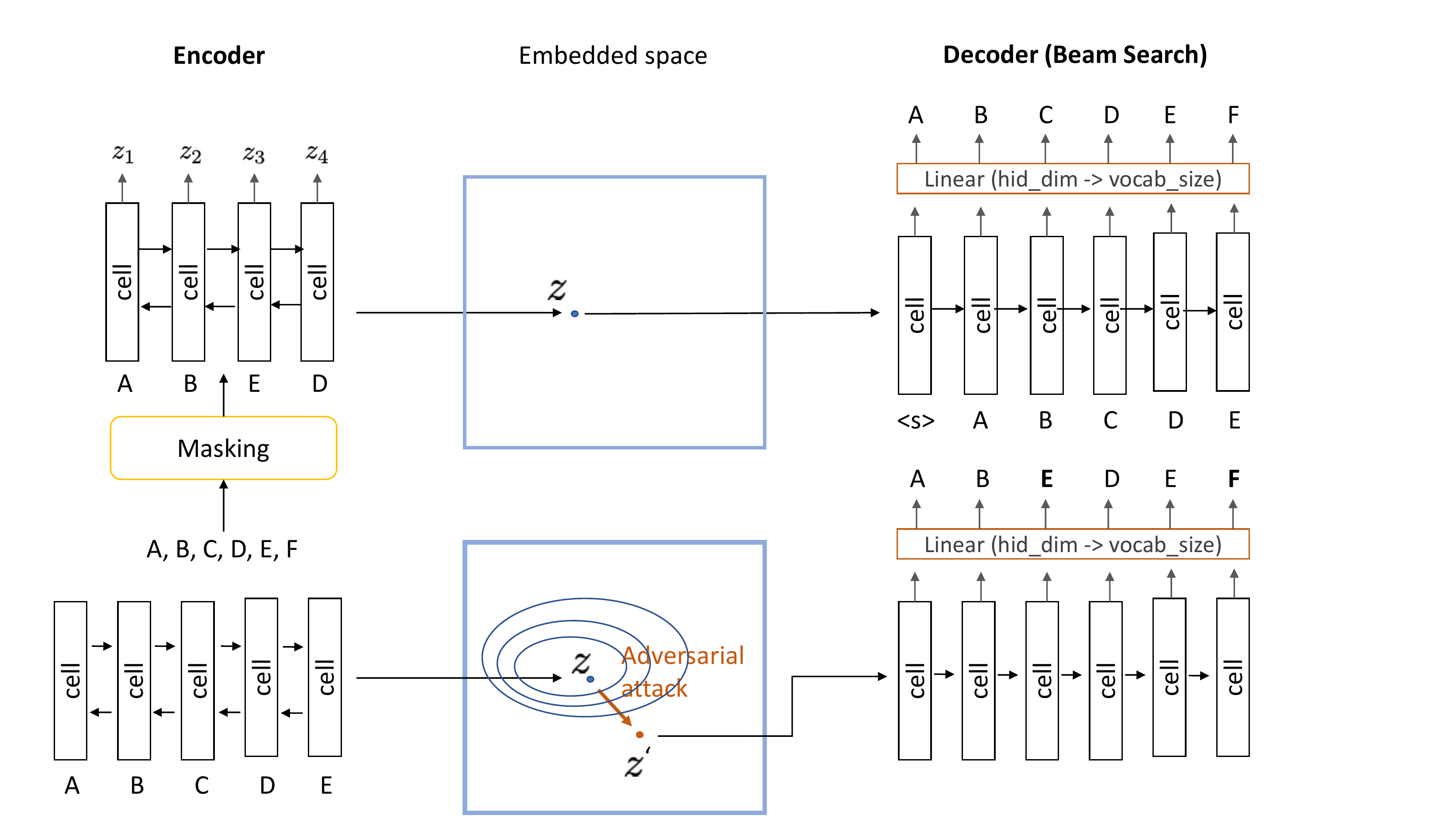}
  \caption{Top figure: learning of our seq2seq model with the masking of tokens in an initial sequence. We also use beam search and an attention mechanism. Bottom figure: our adversarial attack, modification of a sequence $\vecZ$ in the embedded state to be sure that the decoding of the adversarial sequence $D(\vecZ')$ is close to the decoding $D(\vecZ)$, whilst the classifier score is significantly different.}
    \label{fig:teaser}
\end{figure}

\section{Introduction}
\label{sec:intro}

The deep learning revolution has led to the usage of deep neural network-based models across all sectors in the industry: from self-driving cars to oil and gas.
However, the reliability of these solutions are questionable due to the vulnerability of almost all of the deep learning models to adversarial attacks~\cite{yuan2019adversarial} in computer vision~\cite{akhtar2018threat,khrulkov2018art}, NLP~\cite{zhang2019adversarial,wang2019survey}, and graphs~\cite{sun2018adversarial}.
The idea of an adversarial attack is to modify an initial object, so the difference is undetectable to a human eye, but fools a target model: a model misclassifies the generated object, whilst for a human it is obvious that the class of the object remains the same~\cite{Kurakin2017}.

For images we can calculate derivatives of the class probabilities with respect to the colour of pixels in an input image.
Thus, moving along this direction we can apply slight alterations to a few pixels, and get a misclassified image, whilst keeping the image almost the same.
For different problem statements attacks can be different, but in general a continuous space of images is rich enough for providing adversarial images.

The situation is different for sequential categorical data due to its discrete nature and thus absence of partial derivatives with respect to the input.
% Whilst a sequence representation can lie in a continuous space, when we go back to the space of sequences, we can move far away from the initial object, rendering partial derivatives useless. 
The space of possible modifications is also limited. For certain problems a malicious user can not modify an object arbitrarily.
For example, whilst trying to increase a credit score we can not remove a transaction from the history available to the bank; we only add another transaction.
Both of these difficulties impose additional challenges for creation of adversarial attacks for categorical sequential data.

A survey on adversarial attacks for sequences~\cite{zhang2019adversarial,wang2019survey} presents a list of possible options to overcome these difficulties. With respect to white-box attacks, there are two main research directions.
Many approaches work with the initial space of tokens as input attempting to modify these sequences of tokens using operations like addition or replacement~\cite{samanta2017towards,liang2017deep,ebrahimi2018hotflip}.
Another idea is to move into an embedded space and leverage on gradients-based approaches in this space~\cite{sato2018interpretable}.
We also note that most of these works focus on text sequence data.

\begin{table*}[h!]
    \caption{Examples of adversarial sequences generated by the baseline HotFlip and our CASCADA approaches for the AG news dataset. HotFlip often selects the same strong word corrupting the sequence semantics and correctness. CASCADA is more ingenious and tries to keep the semantics, whilst sometimes changing the sequence too much.}
    \label{tab:adversarial_examples}
    \centering
    \begin{tabular}{lll}
    \hline
    Initial sequence $x$ & HotFlip adversarial  & CASCADA adversarial \\
    \hline
      jayasuriya hits back 
 & jayasuriya \textcolor{orange}{arafat} back 
 & \textcolor{orange}{snow} hits \textcolor{orange}{over} back  \\
 for sri lanka & for sri lanka & for sri lanka \\
 \rowcolor{Gray}
  determined jones jumps 
 & \textcolor{orange}{arafat} jones jumps  & \textcolor{orange}{ibm music} jumps \\
  \rowcolor{Gray} 
  into finals & into finals & into \textcolor{orange}{match} \\
   tiny memory card for 
& tiny memory card for 
& \textcolor{orange}{artificial} memory card for 
 \\
 mobiles launched & \textcolor{orange}{economy economy} & \textcolor{orange}{hewitt pistons} \\
 \rowcolor{Gray}
 nokia plots enterprise move
& nokia plots \textcolor{orange}{economy economy}
& nokia \textcolor{orange}{steers} enterprise move \\
sony shrinking the ps & 
  sony shrinking \textcolor{orange}{economy} ps & 
  sony blames the \textcolor{orange}{indies cross} \\
  \rowcolor{Gray}
 sackhappy  d  bags bills &
 \textcolor{orange}{google} d bags bills & 
 \textcolor{orange}{textile} d bags bills \\ 
 
 tunisian president ben 
& \textcolor{orange}{nba} president ben 
& \textcolor{orange}{bayern hat toshiba} \\
ali reelected & ali reelected & \textcolor{orange}{got} reelected \\
    \hline     
    \end{tabular}
\end{table*}

We propose two approaches that can alleviate the aforementioned problems with differentiability and a limited space of modification actions, and work in the space of embedded sequences.
The first approach is based on a Monte-Carlo search procedure in an embedded space, treating as the energy the weighted sum of the distance between the initial sequence and the generated one and the difference between the probability scores for them.
The first term keeps two sequences close to each other, whilst the second term identifies our intention to fool the classifier and generate a similar but misclassified example for a particular object.
This approach is universal, as it does not require derivatives for the first and second terms whilst traversing the embedded space.
The number of hyperparameters remains small, and each hyperparameter is interpretable with respect to the problem statement.
The second approach illustrates adopts differentiable versions of sequential distance metrics.
We use a trained differentiable version of the Levenshtein distance~\cite{moon2018multimodal} and a surrogate classifier defined on embeddings of sequences.
In this case our loss is differentiable, and we can adopt any gradient-based adversarial attack.
The two approaches, which we name MCMC and CASCADA attacks, are summarised in Figure~\ref{fig:teaser}.
Examples of generated sequences for the AG News dataset are presented in Table~\ref{tab:adversarial_examples}.

The generative model for adversarial attacks is a seq2seq model with masking~\cite{bowman2016generating}. So, the constructed RNN model can be reused for generating adversarial attacks based on these two approaches and creating adversarial attacks with a target direction as well as training embeddings for sequences.
The validation of our approaches includes testing on diverse datasets from NLP, bank transactions, and medical insurance domains.

To sum up, we consider the problem of adversarial attack generation for categorical sequential data.
The main contributions of this work are the following.
\begin{itemize}
    \item Our first approach is based on an adaptation of Markov Chain Monte Carlo methods.
    \item Our second approach uses a continuous relaxation of the initial problem. This makes it possible to perform a classic gradient-based adversarial attack after applying a few new tricks.
    \item We construct seq2seq models to generate adversarial attacks using an attention mechanism and a beam search, and test the performance for attacking models based on different principles, e.g. logistic regression for TF-IDF features from a diverse set of domains.
    \item Our adversarial attacks outperform the relevant baseline attacks; thus it is possible to construct effective attacks for categorical sequential data.
\end{itemize}

\section{Related work}

There exist adversarial attacks for different types of data.
The most popular targets for adversarial attacks are images~\cite{szegedy2014intriguing,goodfellow2014explaining}, although some work has also been done in areas such as graph data~\cite{zugner2018adversarial} and sequences~\cite{papernot2016crafting}.

It seems that one of the first articles on the generation of adversarial attacks for discrete sequences is~\cite{papernot2016crafting}. 
The authors correctly identify the main challenges for adversarial attacks for discrete sequence models: a discrete space of possible objects and a complex definition of a semantically coherent sequence.
Their approach considers a white-box adversarial attack with a binary classification problem. 
We focus on black-box adversarial attacks for sequences. This problem statement was considered in~\cite{gao2018black,liang2017deep,jin2020is}. 

Extensive search among the space of possible sequences is computationally challenging~\cite{fursov2019sequence}, especially if the inference time for a neural network is significant.
Authors of~\cite{gao2018black} identify certain pairs of tokens and then permute their positions within these pairs, thus working directly on a token level.
Another black-box approach from~\cite{liang2017deep} also performs a search at the token level.

It is also possible to use gradients for embeddings~\cite{sato2018interpretable}.
However, the authors of~\cite{sato2018interpretable} limit directions of perturbations by moving towards another word in an embedded space, and the authors of~\cite{sato2018interpretable,ren2020generating} traverse the embedding space, whilst achieving limited success due to the outdated or complex categorical sequence models.
Also, they consider only general perturbations and only NLP problems, whilst it is important to consider more general types of sequences.

As we see from the current state of the art, there is still a need to identify an effective end2end way to explore the space of categorical sequences for the problem of adversarial attacks generation. Moreover, as most of the applications focus on NLP-related tasks, there is still a room for improvement by widening the scope of application domains for adversarial attacks on categorical sequences. 
Among the methods presented in the literature we highlight HotFlip~\cite{ebrahimi2018hotflip} as the most justified option, so we use compare it with our embeddings-based methods.

% To leverage the embeddings space for the generation of adversarial sequences we need to define a differentiable mapping from the embedded space to the space of sequences. Another simpler solution would be  to propose an effective procedure for traversing through the embedded space.

% Markov chain Monte Carlo (MCMC) approaches provide a good candidate set of solutions for exploration of embedded spaces~\cite{hastings1970monte}. We select an energy function to reflect our views on criteria for good and bad sequences.
% For, example traversing complex spaces via MCMC is possible for the space of deep neural network parameters~\cite{zhang2019cyclical}.

% Another option is to define a differentiable loss function that takes into account the distance between the initial and generated sequences and simultaneously tries to minimise the class score for the generated adversarial example. 
% Authors of~\cite{cheng2018seq2sick} compose a white-box adversarial attack based on this idea, but due to a high divergence between the embedded and categorical sequence spaces, they return to the categorical sequence space at each step, thus making the search ineffective and not end to end.

% For the class score it is relatively easy to define a differentiable version of it, whilst for the distance measure this problem is more complex.
% However, there are differentaible versions of distance measures, e.g. a differentiable BLEU score~\cite{zhukov2017differentiable}, a learned~\cite{moon2018multimodal} or soft~\cite{ofitserov2019soft} Levenshtein edit distance.

\section{Methods}

We start this section with the description of the general sequence-to-sequence model that we use to generate adversarial sequences, with some necessary details on model training and structure. We then describe the classifier model that we fool using our adversarial model. Next, we describe, how our seq2seq model is used to generate adversarial examples and present our MCMC and CASCADA adversarial attacks. Finally, we provide a description of how to obtain a differentiable version of the Levenshtein distance.

\subsection{Models} 
\label{sec:seq2seqmodel}

\paragraph{Sequence-to-sequence models.}
Seq2seq models achieve remarkable results in various NLP problems, e.g. machine translation~\cite{bahdanau2015neural}, text summarisation~\cite{li2017deep}, and question answering~\cite{hu2017learning}.
These models have an encoder-decoder architecture: it maps an initial sequence $\vecX$ to dense representation using an encoder $\vecZ = E(\vecX)$ and then decodes it using a decoder $\vecX' = D(\vecZ)$ back to a sequence. 

Following the ideas from CopyNet~\cite{gu2016incorporating}, we use a seq2seq model with an attention mechanism~\cite{bahdanau2015neural} for the copying problem and train an encoder and a decoder such that $\vecX' \approx \vecX$. 
The final network is not limited to copying the original sequence, but also discovers the nature of the data providing a language model. 
As the encoder $E(\vecX)$ we use a bi-directional LSTM ~\cite{gers1999learning}, and as the decoder $D(\vecX)$ we use a uni-directional LSTM with Beam Search~\cite{graves2012sequence}.

To train the model we mask some tokens from an input sequence, whilst trying to recover a complete output sequence,
adopting ideas from MASS~\cite{song2019mass} and training a CopyNet~\cite{gu2016incorporating} with the task to reconstruct an initial sequence. 
Masking techniques include swap of two random tokens, random deletion, random replacement by any other token, and random insertion.
The objective for training the model is cross-entropy~\cite{papineni2002bleu}. As we do not need any labelling, this unsupervised problem is easy to define and train.

In addition, we input a set of possible masking operations $\mathbf{m} = \{m_1, \ldots, m_s \}$. 
An example of such a set is $\mathbf{m} = \{AddToken, Replace, Delete\}$. We provide $\mathbf{m}$ to the model in addition to input sequence $\vecX$.
As another example, for bank transactions, we can only use the addition of new tokens and $\mathbf{m} = \{AddToken\}$.

% , as
% this can be the only possible way to modify a sequence of transactions, or add new tokens to the end of a sequence.

\paragraph{Classification models.}

As a classifier $C(\vecX)$ we use a one-layer bi-directional LSTM with one fully-connected layer over the concatenation of the mean $\frac{1}{d} \sum_{i = 1}^d z_i$ and $\max(\vecZ)$ of a hidden state $\vecZ = \{z_1, \ldots, z_d\}$ or a logistic regression with TF-IDF features.
A classifier takes a sequence $\vecX$ as input and outputs class probabilities (a classifier score) $C(\vecX) \in [0, 1]^k$, where $k$ is the number of classes or a class label $c(\vecX)$ on the base of class probability scores $C(\vecX)$.

\subsection{Generation of adversarial sequences} 

We generate adversarial sequences for a sequence $\vecX$ by a targeted modification of a hidden representation $\vecZ = E(\vecX)$ given by encoder $E(\cdot)$ in such a way that the decoder generates an adversarial sequence $A(\vecX)$ that is \begin{inparaenum}[\bfseries(1)] \item similar to the original sequence and \item have a lower probability of a targeted label. \end{inparaenum}

The general attack scheme is presented in Algorithm~\ref{alg:attack}.
This attack works under the black-box settings: an attacker has no access to the targeted model.
The algorithm uses an encoder, a decoder, word error rate $WER$ between a generated and the initial sequences and a classifier that outputs class probability $C(\vecX)$, and a class label $c(\vecX)$.
Slightly abusing the notation we refer to $C = C(\vecX)$ as the classifier score for a class we want to attack in case of multiclass classification.
CASCADA attack also uses a surrogate classifier and a surrogate word error rate distance.

The attack algorithm generates a set $\{\vecZ_1, \ldots, \vecZ_N\}$ of adversarial candidates via consecutive steps $\vecZ_i := G(\vecZ_{i - 1})$ in the embedded space starting at $\vecZ$ and selects the best one from the set.
The difference between algorithms is in which function $G(\vecZ)$ we use.
%at each iteration we use the attack $G(\vecZ)$ as a function that acts in an embedded space to create a diverse set of adversarial candidates.

\begin{minipage}{0.46\textwidth}
\begin{algorithm}[H]
 \KwIn{Number of steps $N$}
 \KwData{Original sequence $\vecX$ and true label $c_\vecX$}
 \KwResult{Adversarial sequence $\vecX^* = A(\vecX)$}
 $\vecZ_0 = E(\vecX)$\;
 \For{$i \gets 1$ \textbf{to} $N$} {
  \%  attack generator step\;
  $\vecZ_i := G(\vecZ_{i - 1})$\;
  $C_i := C(D(\vecZ))$ \% score\;
  generate class label $c_i$ from score $C_i$\;
  $w_i = WER(D(\vecZ_i), \vecX)$\;
 }
  \eIf{$\exists i$ s.t. $c_i \ne c_{\vecX}$}{
   $\vecX^* = \vecX_i$ s.t. $i = \mathrm{arg}\min_{i: c_i \ne c_{\vecX}} w_i$\;
   }
   {
   $\vecX^* = \vecX_i$ s.t. $i = \mathrm{arg}\min_{i} C_i$\;
   }
 \caption{The general attack scheme}
 \label{alg:attack}
\end{algorithm}
\end{minipage}
\hfill
\begin{minipage}{0.46\textwidth}
\begin{algorithm}[H]
 \KwIn{Embedding $\vecZ$, proposal variance $\sigma^2$,
 energy temperatures $\sigma_{wer}$, $\sigma_{class}$, initial class label $c_0$}
 \KwResult{Attacked embedding $\vecZ' = G(\vecZ)$}
  $\boldsymbol{\varepsilon} \sim \mathcal{N}(\mathbf{0}, \sigma^2 I)$\;
  $\vecZ' := \vecZ + \boldsymbol{\varepsilon}$\;
  $\vecX' := D(\vecZ')$\;
  $C: = C(\vecX')$\;
  generate class label $c$ from score $C$\;
  $w = WER(\vecX', \vecX)$\;
  $\alpha = \exp \left( \frac{-w}{\sigma_{wer}} + \frac{-[c_0 = c]}{\sigma_{class}} \right) $\;
   $u \sim \mathcal{U}([0, 1])$\;
   \If{$\alpha < u$}{
   $\vecZ' := \vecZ$\;}  
 \caption{The MCMC attack defines a generator step $\vecZ_i := G(\vecZ_{i - 1})$, $[\cdot]$ is the indicator function}
  \label{alg:mcmc}
\end{algorithm}
\end{minipage}

\subsubsection{Naïve random walk attack.}
The natural approach for generating a new sequence $\vecX^*$ in an embedded space is a random jump to a point $\vecZ^*$ in that embedded space from the embedding of an initial sequence $\vecZ = E(\vecX)$.
An adversarial candidate is a decoder output $\vecX^* = D(\vecZ^*)$.
As we have a total budget $N$, we make up to $N$ steps until we find a sufficiently good sequence.
Whilst this algorithm seems to be quite simple, it can provide a good baseline against more sophisticated approaches, and can work well enough for an adequate embedding space.

Formally, for this variation of Algorithm \ref{alg:attack} we use $\vecZ' = G(\vecZ) = \vecZ + \boldsymbol{\varepsilon}, \boldsymbol{\varepsilon} \sim \mathcal{N}(\mathbf{0}, \sigma^2 I)$ with $\sigma^2$ being a hyperparameter of our algorithm.
Note that in the case of a random walk we defer from the general attack scheme, and each time use the same initial sequence $\vecZ_0 = E(\vecX)$ instead of $\vecZ_{i - 1}$ to get a new sequence~$\vecZ_i$.

%(Algorithm \ref{alg:mcmc})

\subsubsection{MCMC walk.}
Markov chain Monte Carlo (MCMC) can lead to a more effective approach. 
We generate a new point using Algorithm \ref{alg:attack} with $G(\cdot)$ defined in Algorithm \ref{alg:mcmc} by an MCMC walk.
This walk takes into account the similarity between the initial and the generated sequences and the adversity of the target sequence, so we can generate point $\vecZ_i := G(\vecZ_{i - 1})$ at each step more effectively.
Similar to the naïve random walk, the MCMC uses the noise variance for embedded space $\sigma$.
In addition, the MCMC walk approach has temperature parameters $\sigma_{wer}$ and $\sigma_{class}$ that identify the scale of the energy we are seeking, and what is the trade-off between the distance among sequences and the drop in the classification score.

The MCMC random walk is designed to make smarter steps and traverses through the embedded space.

\subsubsection{CASCADA attack.}
\label{sec:cascada_attacl}
Naïve and MCMC attacks can be inefficient. 
Both of these approaches are computationally expensive for deep seq2seq architectures.
%Each step requires a beam search run and a calculation of WER.

The CASCADA (CAtegorical Sequences Continuous ADversarial Attack) attack is an end-to-end approach, which computes the $WER$ metric and runs a beam search only once.

In the CASCADA approach we use Deep Levenshtein model $WER_{deep}(\vecZ, \vecZ')$~\cite{moon2018multimodal} and a surrogate classification model $C_s(\vecZ)$ on top of a seq2seq CopyNet.
Both of these models act in the embeddings space.
Therefore, we can evaluate derivatives with respect to arguments of $WER_{deep}(\vecZ_0, \vecZ)$ and $C_s(\vecZ)$ inside the target function, thus making it possible to run a gradient-based optimisation that tries to select the adversarial sequence with the best score.

We search for a minimum of a function $C_s(\vecZ) + \lambda WER_{deep}(\vecZ, \vecZ_0)$ with respect to $\vecZ$.
The hyperparameter $\lambda$ identifies a trade-off between trying to get a lower score for a classifier and minimising the distance between $\vecZ$ and the initial sequence $\vecZ_0$.
So, the attack $\vecZ'$ is a solution of the optimisation problem:
\[
\vecZ' = \mathrm{arg}\min_{\vecZ} C_s(\vecZ) + \lambda WER_{deep}(\vecZ, \vecZ_0).
\]

After the generation of a set of candidates during the gradient descent optimisation $\vecZ_1, \ldots, \vecZ_N$, we apply the decoder to each candidate, obtaining $\vecX_1 = D(\vecZ_1), \ldots, \vecX_N = D(\vecZ_N)$ as a set of adversarial candidates.

\paragraph{Deep Levenshtein.}
To make gradient-based updates to an embedded state, we use a differentiable version of the Levenshtein distance function~\cite{fursov2020differentiable}. 
We use the Deep Levenshtein distance proposed by \cite{moon2018multimodal} and considered also in~\cite{fursov2020differentiable}.
In our case, $WER$ is used instead of the Levenshtein distance, since we work on the word level instead of the character level for NLP tasks, and for non-textual tasks there are simply no levels other than ``token'' level.

To collect the training data for each dataset we generate about 2 million pairs.
For each pair we apply masks similar to CopyNet, obtaining an original sequence and a close but different sequence. 
We have also added pairs composed of different sequences from the training data 
for a better coverage of distant sequences.
Our target is $WER_{norm}(\vecX, \vecY) = \frac{WER(\vecX, \vecY)}{\max(|\vecX|, |\vecY|)}$. We train a model $M(\vecZ)$ with the objective $\|\frac12 (\cos(M(E(\vecX)), M(E(\vecY))) + 1) - WER_{norm}(\vecX, \vecY) \|$.
% To improve the quality of the model, we concatenate $M(\vecX)$ and corresponding attention vectors of $M(\vecY)$.
The mean absolute error for the learned Deep Levenstein distance $WER_{deep}(\vecZ, \vecZ') = \frac12 (\cos(M(\vecZ), M(\vecZ')) + 1)$ is $0.15$ for all considered datasets.

\vspace{-3mm}
\section{Experiments}
\vspace{-3mm}
In this section we describe our experiments. The datasets and the source code are published online\footnote{The code is available at \\ \url{https://github.com/fursovia/dilma/tree/master}. The data is available at \url{https://www.dropbox.com/s/axu26guw2a0mwos/adat_datasets.zip?dl=0}.}.
\vspace{-3mm}

\subsection{Datasets}
\vspace{-1mm}
To test the proposed approaches we use NLP, bank transactions, and medical sequence datasets.

% \cite{hu2017toward}
We use \textbf{NLP dataset} AG news~\cite{zhang2015character} dedicated to topic identification.
The four largest classes from the corpus constitute our dataset. The number of training
samples for each class is $30,000$ and the number of test samples is $1,900$. 
We also use \textbf{a transactions dataset}, aimed at predicting gender~\footnote{\url{https://www.kaggle.com/c/python-and-analyze-data-final-project/data}}.
We use sequences of transactions codes (gas station, art gallery, etc.) and transaction amounts as an input.
% There are two open \textbf{transactions datasets} that we use in our work, aimed at predicting age~\footnote{\url{https://onti.ai-academy.ru/competition}} and gender~\footnote{\url{https://www.kaggle.com/c/python-and-analyze-data-final-project/data}}.
% We use sequences of transactions codes (gas station, art gallery, and so on) and transaction amounts as input for both datasets.
We also supplement these datasets with another dataset from the \textbf{medical insurance}~\cite{fursov2019sequence} domain.
The goal is to detect frauds based on a history of visits of patients to a doctor. Each sequence consists of visits with information about a drug code and amount of money spent for each visit.

For the attacked logistic regression model with TF-IDF features as inputs, the macro-average ROC AUC scores for Transcations-GENDER, Healthcare Insurance and AG News datasets are $0.70$, $0.74$, $0.88$, and $0.96$ correspondingly. % Transactions-AGE

\textbf{Preprocessing of the datasets.}
%\label{sec:preprocess}
For AG news we use a standard preprocessing procedure.
For the healthcare insurance dataset each sequence of tokens consists of medical codes or the procedure assigned after the next visit to a clinic, and a label if the entire sequence for a patient is a fraud or not, with the percentage of frauds in the available dataset being $1.5 \%$ and total number of patients being $381,013$.

For the transactions datasets the preprocessing is more complex, so we describe it separately.
For the gender prediction dataset we compose each token from the transaction type, the Merchant Category Code (MCC), and the transaction amount bin.
We split all amounts into decile bins and then sort them, so index $0$ corresponds to the cheapest purchases and index $9$ corresponds to the most expensive purchases.
An example encoding of a token from a sequence of transactions is $4814\_1030\_3$ with $4814$ being the MCC code, $1030$ being the transaction type and $3$ the index of the decile amount bin.
Each sequence corresponds to transactions during the last three days with the mean sequence length being $10.25$.
% For the age prediction dataset we proceed in a similar way, but use only the transaction type instead of both the MCC code and the transaction type.

% For the transactions datasets the preprocessing is more complex, so we describe it separately.
% For the gender prediction dataset we compose each token from the transaction type, the Merchant Category Code (MCC), and the transaction amount bin.
% We split all amounts into decile bins and then sort them, so index $0$ corresponds to the cheapest purchases and index $9$ corresponds to the most expensive purchases.
% An example encoding of a token from a sequence of transactions is $4814\_1030\_3$ with $4814$ being the MCC code, $1030$ being the transaction type and $3$ the index of the decile amount bin.
% Each sequence corresponds to transactions during the last three days with the mean sequence length being $10.25$.
% For the age prediction dataset we proceed in a similar way, but use only the transaction type instead of both the MCC code and the transaction type.
%An example encoding of a token from such this sequence is $35\_7$ with $35$ being the transaction type and $7$ the index of the decile amount bin.

% Model qualities for different classification and seq2seq models to motivate the selection of the model described in Section~\ref{sec:seq2seqmodel}.
% We also compare various options for the training of the Deep Levenshtein distance defined in Section~\ref{sec:deep_leve}.

\subsection{Metrics}

The two types of metrics for the evaluation of the quality of adversarial attacks on sequences are the difference in the classifier score between an initial and a generated adversarial sequences and the distance between these sequences.

To measure the performance of the proposed approaches we use three metrics that identify the accuracy drop after adversarial attacks: the ROC AUC drop, the accuracy drop, and the mean classifier score drop.
To measure the difference for the new adversarial sequences we use the word error rate ($WER$) between the initial and generated adversarial sequences.

We also propose a new metric for evaluating adversarial attacks on classifiers for categorical sequences, which combines distance-based and score-based approaches. 
To get a more realistic metric we perform a normalisation using $WER$s between the initial and adversarial sequences, which we call the normalised accuracy drop 
% Misclassification error does not take into account how similar are adversarial and non-adversarial examples. We can fool any symbolic sequence classifier by changing the initial sequence a lot, however, such changes are easier to be caught.
$
\mathrm{NAD}(A) = \frac{1}{|Z|} \sum_{i \in Z} 1\{c(\vecX_i) \neq c(A(\vecX_i)) \} \left(\frac{L_i - \mathrm{WER}(A(\vecX_i), \vecX_i)}{L_i - 1}\right),
$
where $c(\vecX)$ outputs class labels instead of probabilities $C(\vecX)$, $Z=\{i| c(\vecX_i)=y_i\}$, and $L_i$ is the maximum length of $\vecX_i$ and the adversarial sequence $\vecX'_i = A(\vecX_i)$ generated by the adversarial attack $A$.

\subsection{Main experiment for adversarial attacks}

We compare our approach with the current state of the art, HotFlip~\cite{ebrahimi2018hotflip}. HotFlip at each step selects the best token to change, given an approximation of partial derivatives for all tokens and all elements of the dictionary.
To complete the HotFlip attack in our setting we generate $N$ sequences with beam search and then follow our general selection procedure described in Algorithm~\ref{alg:attack}.

We run experiments to keep $WER$ similar for the four considered approaches: HotFlip, random walk attack, MCMC walk attack, and CASCADA.
We select hyperparameters to get approximately similar $WER$  scores for different approaches.
We generate $N = 100$ sequences for each of the four approaches and select the best one according to the criterion described above.

In Table~\ref{table:aa_quality} we present results for the proposed approaches, whilst attacking an independent logistic regression model with TF-IDF features and using LSTM model as a surrogate classifier. 
We see that embedding-based approaches provide decent performance and are a better way to generated more adversarial examples, while NAD metric puts too significant emphasis on $WER$ values when comparing different approaches.
% In Table~\ref{table:aa_quality} we present results for proposed approaches whilst attacking an independent LSTM model with the end to end training of the embeddings. 
% For adversarial attacks we use a seq2seq end to end model based on the LSTM architecture.

\begin{table}[t!]
\centering
\caption{Fooling logistic regression with TF-IDF representations as inputs by running the considered attacks on the four diverse datasets.
We maximise metrics with the $\uparrow$ signs and minimise metrics with the $\downarrow$ signs.
Embedding-based methods work better when looking both at perplexity and accuracy drops.}
\label{table:aa_quality}
\begin{tabular}{ccccccc}
% \hline
% Transactions,  & ROC AUC  & Accuracy & Probability  & Normalised &  Log  & NAD $\uparrow$ \\
% Age & drop $\uparrow$ & drop $\uparrow$ & drop $\uparrow$ & $WER$ $\downarrow$ & perplexity $\downarrow$ & \\
% \hline
% Random walk & 0.229 & {\bf 0.37} & 0.109 & 0.477 & {\bf 3.79} & {\bf 0.493} \\
% HotFlip & 0.020 & 0 & 0.025 & {\bf 0.120} & 4.83 & 0.319 \\
% MCMC walk & 0.065 & 0.09 & 0.034 & 0.162 & 3.84 & 0.129 \\
% CASCADA & {\bf 0.235} & 0.29 & {\bf 0.111} & 0.186 & 4.05 & 0.479 \\
\hline
Transactions & ROC AUC  & Accuracy & Probability  & Normalised &  Log  & NAD $\uparrow$ \\
Gender & drop $\uparrow$ & drop $\uparrow$ & drop $\uparrow$ & $WER$ $\downarrow$ & perplexity $\downarrow$ & \\
\hline
Random walk & 0.539 & 0.40 & 0.189 & 0.561 & 4.29 & 0.334 \\
HotFlip & 0.243 & 0.26 & 0.091 & {\bf 0.100} & 5.15 & {\bf 0.623} \\
MCMC walk & {\bf 0.640} & {\bf 0.55} & {\bf 0.245} & 0.719 & {\bf 4.28} & 0.333 \\
CASCADA & 0.361 & 0.32 & 0.121 & 0.198 & 4.49 & 0.426 \\
\hline
AG News & & & & & & \\
\hline
Random walk & 0.406 & 0.66 & 0.487 & 0.704 & 5.21 & 0.274 \\
HotFlip & 0.342 & 0.67 & 0.477 & {\bf 0.218} & 6.76 & {\bf 0.723} \\
MCMC walk & {\bf 0.452} & {\bf 0.72} & {\bf 0.525} & 0.757 & {\bf 5.16} & 0.270 \\
CASCADA & 0.422 & 0.62 & 0.492 & 0.385 & 6.29 & 0.494 \\
\hline
Healthcare insurance & & & & & & \\
\hline
Random walk & 0.566 & 0.47 & 0.094 & 0.725 & 4.90 & 0.258 \\
HotFlip & {\bf 0.778} & {\bf 0.92} & {\bf 0.294} & {\bf 0.464} & 6.75 & {\bf 0.371} \\
MCMC walk & 0.364 & 0.29 & 0.062 & 0.695 & 4.50 & 0.194 \\
CASCADA & 0.131 & 0.26 & 0.045 & 0.492 & {\bf 4.28} & 0.106 \\
\hline
\end{tabular}
\end{table}

\begin{table}[t!]
    \centering
        \caption{Constrained adversarial attacks on logistic regression with TF-IDF using various masking tokens for the AG news dataset. Log perplexity is almost similar for all approaches.}
    \label{table:maskers}
\begin{tabular}{lcccc}
\hline
Masker & Accuracy & Normalised & NAD $\uparrow$  \\
 & drop $\uparrow$ & $WER$ $\downarrow$ & \\
\hline
No constraints & {\bf 0.62} & {\bf 0.39} & {\bf 0.492}  \\
Add & {\bf 0.62} & 0.51 & 0.382  \\
Replace & 0.59 & 0.50 & 0.366 \\
Swap & 0.61 & 0.52 & 0.333  \\
    \hline
    \end{tabular}
\end{table}

\subsection{Constrained adversarial attack}

We compare the performance of general and constrained adversarial attacks.
In the first case the attack applies all possible modifications to sequences.
In the second case only certain perturbations are allowed, e.g. an addition of a token or swapping two tokens.
The comparison of performances for various attacks is presented in Table~\ref{table:maskers}: all types of attacks have comparable performances for our CASCADA approach.

\subsection{Reliability study}

\begin{figure}
    \centering
    \includegraphics[width=0.8\textwidth]{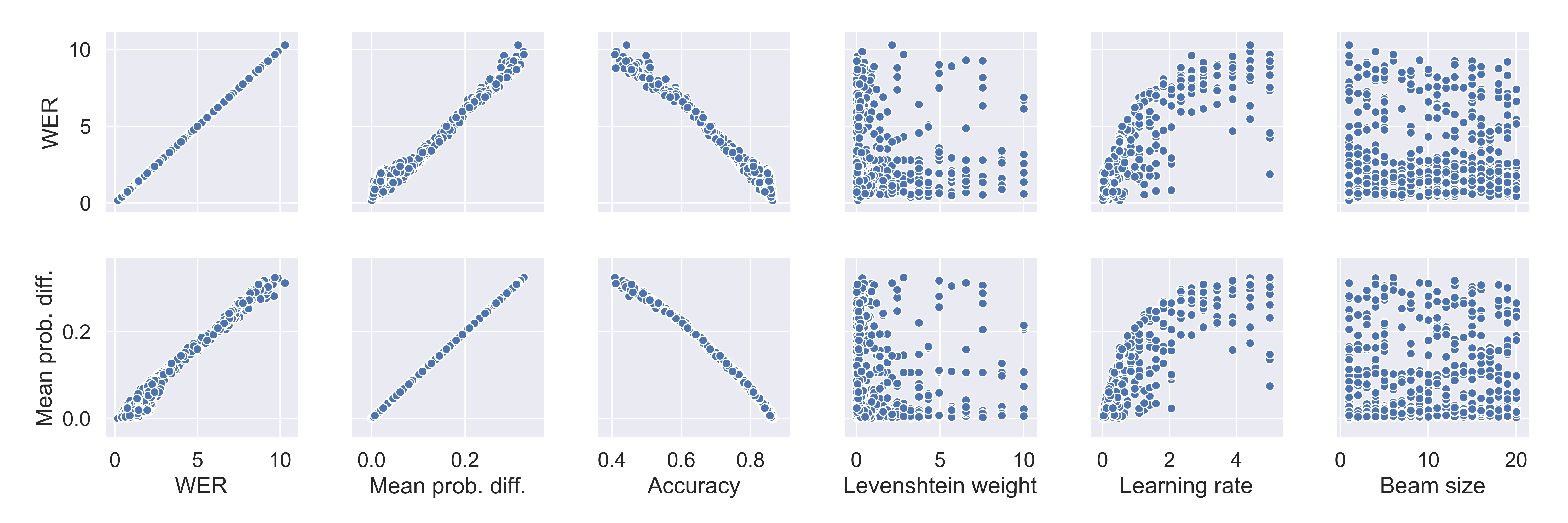}
    \caption{Mean $WER$ and accuracy drops for various configurations of hyperparameters for the Transactions Gender
dataset: the learning rate, the Deep Levenshtein weight, and the beam number. Mean $WER$ and accuracy drop are
inversely related as expected, whilst the seq2seq model is robust against changes of hyperparameter values.}
    \label{fig:reliability}
\end{figure}

The selection of hyperparameters often affects the performance of an adversarial attack. 
We run 599 different hyperparameters configurations 
for training seq2seq models, trained with attention and
masking, and the CASCADA adversarial attack based on
this model. 
The results are presented in Figure~\ref{fig:reliability}.
We observe that by varying hyperparameters, we select a trade-off between the similarity of initial sequence and an adversarial one and corresponding classifier probability drop.
Moreover, varying of hyperparameters for a selected trade-off we observe robust results without significant drop of quality for particular runs or particular configurations.

% We observe that the mean values of $WER$ and the accuracy drop are inversely related for all considered hyperparameters: there is no silver bullet that provides an adversarial sequence for the initial sequence with corrupted
% probabilities for the true class, which is close to the
% initial sequence.

\vspace{-3mm}
\section{Conclusion}
\vspace{-3mm}
A construction of an adversarial attack for a categorical sequence is a challenging problem. We consider two approaches to solve this problem: directed random modifications and two differentiable surrogates, for a distance between sequences and for a classifier, that act from an embedded space. The first approach is based on the application of MCMC to generated sequences, and the second approach uses surrogates for constructing gradient attacks.
At the core of our approaches lies a modern seq2seq architecture, which  demonstrates an adequate performance. To improve results we adopt recent ideas from the NLP world, including masked training and the attention mechanism. 

For considered applications, which include NLP, bank card transactions, and healthcare, our approaches show a reasonable performance with respect to common metrics for adversarial attacks and sequence distances. Moreover, we can limit the space of possible modifications, e.g. use only addition operations during an adversarial sequence generation.
\vspace{-4mm}
\section*{Acknowledgments}
\vspace{-3mm}
The work of Alexey Zaytsev was supported by  RSF grant 20-71-10135. The work of Evgeny Burnaev was supported by RFBR grant 20-01-00203.
\vspace{-3mm}

{\footnotesize
\bibliographystyle{ieeetr}
\bibliography{adversarial}
}

\end{document}